\documentclass[preprint,12pt]{elsarticle}

\usepackage{amssymb}
\usepackage{multirow}
\usepackage{booktabs}
\usepackage{mathrsfs}
\usepackage{longtable}
\usepackage{lscape}
\usepackage{tabularx}
\usepackage{epsfig}
\usepackage{colortbl}
\usepackage{slashbox}
\usepackage{dcolumn,longtable,hhline,colortbl}
\usepackage[table]{xcolor}
\usepackage[rotateright]{rotating}
\usepackage{multirow}
\usepackage{morefloats}
\usepackage{amsthm}

\newdefinition{definition}{Definition}

\journal{Expert Systems with Applications}

\begin{document}

\begin{frontmatter}



\title{Distance function of D numbers}


\author[swu]{Meizhu Li}
\author[swu]{Qi Zhang}
\author[swu]{Xinyang Deng}
\author[swu,vu2]{Yong Deng\corref{cor}}
\ead{prof.deng@hotmail.com; ydeng@swu.edu.cn}

\cortext[cor]{Corresponding author: Yong Deng, School of Computer
and Information Science, Southwest University, Chongqing 400715,
China. TEL: +86-23-68254555}

\address[swu]{School of Computer and Information Science, Southwest University, Chongqing, 400715, China}
\address[vu2]{School of Engineering, Vanderbilt University, Nashville, TN, 37235, USA}

\begin{abstract}
Dempster-Shafer theory is widely applied in uncertainty modelling and knowledge reasoning due to its ability of expressing uncertain information. A distance between two basic probability assignments(BPAs) presents a measure of performance for identification algorithms based on the evidential theory of Dempster-Shafer. However, some conditions lead to limitations in practical application for Dempster-Shafer theory, such as exclusiveness hypothesis and completeness constraint. To overcome these shortcomings, a novel theory called D numbers theory is proposed. A distance function of D numbers is proposed to measure the distance between two D numbers. The distance function of D numbers is an generalization of distance between two BPAs, which inherits the advantage of Dempster-Shafer theory and strengthens the capability of uncertainty modeling. An illustrative case is provided to demonstrate the effectiveness of the proposed function.

\end{abstract}

\end{frontmatter}

\section{Introduction}\label{Introduction}
Dempster-Shafer theory of evidence \cite{dempster1967upper,shafer1976mathematical}, also called Dempster-Shafer theory or evidence theory, is used to deal with uncertain information. This theory needs weaker conditions than bayesian theory of probability, so it is often regarded as an extension of the bayesian theory \cite{dempster1967generalization}. As an effective theory of evidential reasoning, Dempster-Shafer theory has an advantage of directly expressing various uncertainties, so it has been widely used in many fields \cite{bloch1996some,srivastava2003applications,cuzzolin2008geometric,masson2008ecm,denoeux2011maximum,Deng2011,Dengmodeling2011,Dengnew2011,Zhang2012,Kang2012,denoeux2013maximum,yang2013discounted,yang2013novel,wei2013identifying,liu2013evidential,deng2014supplier}. Due to improve the Dempster-Shafer theory of evidence, many studies have been devoted for combination rule of evidence \cite{yager1996aggregation,gebhardt1998parallel,yang2013discounted,lefevre2013preserve,yang2013evidential}, confliction problem \cite{yager1987dempster,lefevre2002belief,liu2006analyzing,schubert2011conflict,tchamova2012behavior}, generation of mass function \cite{bastian2010universal,cappellari2012systematic,liu2013belief,xu2013new,burger2013randomly,liu2014credal}, uncertain measure of evidence \cite{klir1991generalized,bachmann2010uncertainty,bronevich2010measures,baker2012measuring}, and so on \cite{couso2010independence,limbourg2010uncertainty,jirouvsek2011compositional,luo2012agent,karahan2013persistence,mao2014model,zhang2014response}.

Though the Dempster-Shafer theory has an advantage of directly expressing the \lq \lq uncertainty \rq \rq, by assigning the probability to the subsets of the set composed of multiple objects, rather than to each of the individual objects. However, the mathematical framework of Dempster-Shafer theory is based on some strong hypotheses regarding the frame of discernment and basic probability assignment, which limit the ability of Dempster-Shafer theory to represent information in other situations. One of the hypotheses is that the elements in the frame of discernment are required to be mutually exclusive. In many situations, this hypothesis is difficult to satisfied. For example, linguistic assessments shown as \lq \lq Very Good\rq \rq, \lq \lq Good\rq \rq,\lq \lq Fair\rq \rq,\lq \lq Bad\rq \rq,\lq \lq Very Bad\rq \rq. Due to these assessments is based on human judgment, they inevitably contain intersections \cite{liu2012new,zhang2013ifsjsp}. The exclusiveness between these propositions can't be guaranteed precisely, so that the Dempster-Shafer theory is not reasonable for this situation. To overcome the existing shortcomings in Dempster-Shafer theory, a new representation of uncertain information is proposed, which is called D numbers \cite{Deng2012DNumbers,Deng2014EnvironmentDNs,Deng2014DAHPSupplier,Deng2014BridgeDNs,deng2014d}.

Due to present the measure of performance for identification algorithms based on Dempster-Shafer theory, the concept of distance between BPAs has been proposed before \cite{tessem1993approximations,zouhal1998evidence,bauer1997approximation,jousselme2001new,jousselme2012distances,huang2013new}. In order to express the distance between two D numbers, a distance function of D numbers is proposed in this paper. The proposed distance function of D numbers is an extension for the distance function between two BPAs, which is proposed by Anne-Laure Jousselme \cite{jousselme2001new}. In the distance function of D numbers, the frame of discernment are not required to be mutually exclusive. In the situation that the discernment is mutually exclusive, the proposed distance function of D numbers is degenerated as the distance function between two BPAs.

The rest of this paper is organized as follows. Section \ref{Preliminaries} introduces some basic concepts about the Dempster-Shafer theory and the distance between two BPAs. In section \ref{Proposed function} the proposed distance function based on D numbers is presented. Section \ref{Case study} uses an example to compare the differences between the distance of BPAs and the distance of the D numbers. Conclusion is given in Section \ref{Conclusion}.

\section{Preliminaries}\label{Preliminaries}

\subsection{Dempster-Shafer theory of evidence}

Dempster-Shafer theory of evidence \cite{dempster1967upper,shafer1976mathematical}, also called Dempster-Shafer theory or evidence theory, is used to deal with uncertain information. As an effective theory of evidential reasoning, Dempster-Shafer theory has an advantage of directly expressing various uncertainties. This theory needs weaker conditions than bayesian theory of probability, so it is often regarded as an extension of the bayesian theory. For completeness of the explanation, a few basic concepts are introduced as follows.

\begin{definition}
Let $\Omega$ be a set of mutually exclusive and collectively
exhaustive, indicted by
\begin{equation}
\Omega  = \{ E_1 ,E_2 , \cdots ,E_i , \cdots ,E_N \}
\end{equation}
The set $\Omega$ is called frame of discernment. The power set of
$\Omega$ is indicated by $2^\Omega$, where
\begin{equation}
2^\Omega   = \{ \emptyset ,\{ E_1 \} , \cdots ,\{ E_N \} ,\{ E_1
,E_2 \} , \cdots ,\{ E_1 ,E_2 , \cdots ,E_i \} , \cdots ,\Omega \}
\end{equation}
If $A \in 2^\Omega$, $A$ is called a proposition.
\end{definition}

\begin{definition}
For a frame of discernment $\Omega$,  a mass function is a mapping
$m$ from  $2^\Omega$ to $[0,1]$, formally defined by:
\begin{equation}
m: \quad 2^\Omega \to [0,1]
\end{equation}
which satisfies the following condition:
\begin{eqnarray}
m(\emptyset ) = 0 \quad and \quad \sum\limits_{A \in 2^\Omega }
{m(A) = 1}
\end{eqnarray}
\end{definition}

In Dempster-Shafer theory, a mass function is also called a basic
probability assignment (BPA). If $m(A) > 0$, $A$ is called a focal
element, the union of all focal elements is called the core of the
mass function.

\begin{definition}
For a proposition $A \subseteq \Omega$, the belief function
$Bel:\;2^\Omega   \to [0,1]$ is defined as
\begin{equation}
Bel(A) = \sum\limits_{B \subseteq A} {m(B)}
\end{equation}
The plausibility function $Pl:\;2^\Omega   \to [0,1]$ is defined
as
\begin{equation}
Pl(A) = 1 - Bel(\bar A) = \sum\limits_{B \cap A \ne \emptyset }
{m(B)}
\end{equation}
where $\bar A = \Omega  - A$.
\end{definition}

Obviously, $Pl(A) \ge Bel(A)$, these functions $Bel$ and $Pl$ are
the lower limit function and upper limit function of proposition
$A$, respectively.

Consider two pieces of evidence indicated by two BPAs $m_1$ and
$m_2$ on the frame of discernment $\Omega$, Dempster's rule of
combination is used to combine them. This rule assumes that these
BPAs are independent.

\begin{definition}
Dempster's rule of combination, also called orthogonal sum,
denoted by $m = m_1 \oplus m_2$, is defined as follows
\begin{equation}
m(A) = \left\{ {\begin{array}{*{20}l}
   {\frac{1}{{1 - K}}\sum\limits_{B \cap C = A} {m_1 (B)m_2 (C)} \;,} & {A \ne \emptyset ;}  \\
   {0\;,} & {A = \emptyset }.  \\
\end{array}} \right.
\end{equation}
with
\begin{equation} \label{q_8}
K = \sum\limits_{B \cap C = \emptyset } {m_1 (B)m_2 (C)}
\end{equation}
\end{definition}

where $B$ and $C$ are also elements of $2U$, and K is a constant to show the conflict between
the two BPAs.

Note that the Dempster's rule of combination is only applicable to
such two BPAs which satisfy the condition $K < 1$.

\subsection{Distance between two BPAs}

In \cite{jousselme2001new}, a method for measuring the distance between two basic probability assignments is proposed. This distance is defined as follows.

\begin{definition}
Let $m_1$ and $m_2$ be two BPAs on the same frame of discernment $\Omega$, containing $N$ mutually exclusive and exhaustive hypotheses. The distance between $m_1$ and $m_2$ is:

\begin{equation} \label{q_9}
{d_{BPA}}({m_1},{m_2}) = \sqrt {{\frac{1}{2}{(\vec {{m_1}}  - \vec {{m_2}} )}^T}\underline{\underline D} (\vec {{m_1}}  - \vec {{m_2}} )}
\end{equation}

where $\underline{\underline D}$  is a $({2^\Omega } \times {2^\Omega })$-dimensional matrix with $\underline{\underline D} ({A},{B}) = \frac{{\left| {{A} \cap {B}} \right|}}{{\left| {{A} \cup {B}} \right|}}$, and $A \in {2^\Omega }$, $B \in {2^\Omega }$ are the names of columns and rows respectively, note that ${{\left| {\phi  \cap \phi } \right|} \mathord{\left/
 {\vphantom {{\left| {\phi  \cap \phi } \right|} {\left| {\phi  \cup \phi } \right|}}} \right.
 \kern-\nulldelimiterspace} {\left| {\phi  \cup \phi } \right|}} = 0$. Given a bpa $m$ on frame $\Omega$, $\vec {m}$ is a ${2^\Omega }$-dimensional column vector (can also be called a ${2^\Omega } \times 1$ matrix) with ${m_{A \in {2^\Omega }}}(A)$ as its ${2^\Omega }$ coordinates.

 $(\vec {{m_1}}  - \vec {{m_2}})$ stands for vector subtraction and ${(\overrightarrow m )^T}$ is the transpose of vector (or matrix) $\vec {{m_1}}$. When $\vec {{m_1}}$ is a ${2^\Omega }$-dimensional column vector, ${(\overrightarrow m )^T}$ is its ${2^\Omega }$-dimensional row vector with the same coordinates.

\end{definition}

From Definition $5$, another way to write $d_{BPA}$ is:

\begin{equation} \label{q_10}
{d_{BPA}}({m_1},{m_2}) = \sqrt {\frac{1}{2}({{\left\| {\vec {{m_1}} } \right\|}^2} + {{\left\| {\vec {{m_2}} } \right\|}^2} - 2\left\langle {\vec {{m_1}} ,\vec {{m_2}} } \right\rangle )}
\end{equation}

where ${\left\langle {\vec {{m_1}} ,\vec {{m_2}} } \right\rangle }$ is the scalar product defined by

\begin{equation} \label{q_11}
\left\langle {\vec {{m_1}} ,\vec {{m_2}} } \right\rangle  = \sum\limits_{i = 1}^{{2^\Omega }} {\sum\limits_{j = 1}^{{2^\Omega }} {{m_1}({A_i}){m_2}({A_j})\frac{{\left| {{A_i} \cap {A_j}} \right|}}{{\left| {{A_i} \cup {A_j}} \right|}}} }
\end{equation}

with ${A_i},{A_j} \in P(\Theta )$ for $i,j=1,...{2^\Omega }$. ${\left\| {\vec m } \right\|^2}$ is then the square norm of ${\vec m }$:
\begin{equation} \label{q_12}
{\left\| {\vec m } \right\|^2} = \left\langle {\vec m ,\vec m } \right\rangle
\end{equation}

\section{New distance function based on D numbers}\label{Proposed function}
\subsection{D numbers}
In the mathematical framework of Dempster-Shafer theory, the basic probability assignment(BPA) defined on the frame of discernment is used to express the uncertainty quantitatively. The framework is based on some strong hypotheses, which limit the Dempster-Shafer theory to represent some types of information.

One of the hypotheses is that the elements in the frame of discernment are required to be mutually exclusive. In many situations, this hypothesis is difficult to satisfied. For example, linguistic assessments shown as \lq \lq Very Good\rq \rq, \lq \lq Good\rq \rq,\lq \lq Fair\rq \rq,\lq \lq Bad\rq \rq,\lq \lq Very Bad\rq \rq. Due to these assessments is based on human judgment, they inevitably contain intersections. The exclusiveness between these propositions can't be guaranteed precisely, so that the Dempster-Shafer theory is not reasonable for this situation.

To overcome the existing shortcomings in Dempster-Shafer theory, a new representation of uncertain information called D numbers \cite{Deng2012DNumbers} is defined as follows.
\begin{definition}
Let $\Omega$ be a finite nonempty set, a D number is a mapping formulated by
\begin{equation}\label{q_13}
  d:\Omega\rightarrow[0,1]
\end{equation}
with
\begin{equation}\label{q_14}
  \sum_{B\subseteq\Omega}{d(B)\leq1} \quad  and \quad d(\phi)=0
\end{equation}
\end{definition}

where $\phi $ is an empty set and $B$ is a subset of $\Omega$.

It seems that the definition of D numbers is similar to the definition of BPA. But note that, the first difference is the concept of the frame of discernment in Dempster-Shafer evidence theory. The elements in the frame of discernment $\Omega$ of D numbers do not require mutually exclusive. Second, the completeness constraint is released in D numbers. If $\sum_{B\subseteq\Omega}{d(B)=1}$, the information is said to be complete; if $\sum_{B\subseteq\Omega}{d(B)\leq1}$, the information is said to be incomplete.

\subsection{Relative matrix and intersection matrix}
In this paper, a new distance function based on D numbers is proposed to measure the distance between two D numbers. For the frame of discernment is not required to be a mutually exclusive and collectively exhaustive set in D numbers theory, a relative matrix is needed to express the relationship between every D numbers. The relation matrix are defined as follows.

\begin{definition}
Let the number $i$ and number $j$ of n linguistic constants are expressed by $L_i$ and $L_j$, the intersection area between $L_i$ and $L_j$ is $S_{ij}$, and the union area between $L_i$ and $L_j$ is $U_{12}$. The non-exclusive degree $E_{ij}$ can be defined as follows:

\begin{equation}\label{q_15}
  E_{ij}=\frac{S_{ij}}{U_{ij}}
\end{equation}

A relative Matrix for these elements based on the non-exclusive degree can be build as below:
\begin{equation}\label{q_16}
  R=
\left[
\begin{array}{ccccccc}
      &  L_1     &  L_2    & \ldots  &  L_i    &\ldots & L_n\\
 L_1  & 1        &  E_{12} &\ldots  &  E_{1i} &\ldots  &  E_{1n}  \\
 L_2  &  E_{21}  &  1      &\ldots  &  E_{2i} &\ldots  &  E_{2n}  \\
 \vdots  & \vdots  & \vdots &\ldots      &\vdots     &\ldots&  \vdots  \\
 L_i  &  E_{i1}  &  E_{i2} &\ldots  &  1 &\ldots &  E_{in}\\
 \vdots  & \vdots  & \vdots &\ldots      &\vdots     &\ldots&  \vdots  \\
 L_n  &  E_{n1}  &  E_{n2} &\ldots  &  E_{ni} &\ldots & 1
\end{array}
\right]
\end{equation}

\end{definition}

\begin{figure}[htbp]
\begin{center}
\psfig{file=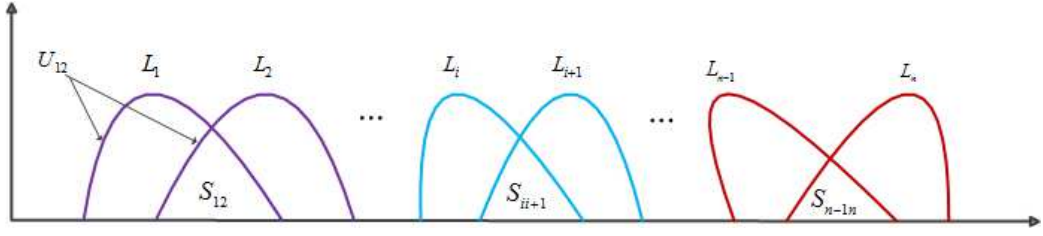,scale=0.75} \caption{Example for linguistic constants} \label{linguistic constants}
\end{center}
\end{figure}

For example, assume $n$ linguistic constants are illustrated as the Fig. \ref{linguistic constants} shows. Based on the area of intersection $S_{ij}$ and union $U_{ij}$  between any two linguistic constants $L_i$ and $L_j$, the non-exclusive degree $E_{ij}$ can be calculated to represent the non-exclusive degree between two D numbers.

\begin{definition}
Given an intersection matrix $\underline{\underline  I}$ between two subsets belong to ${2^\Omega }$. The intersection degree of two subsets, and the intersection degree can be defined as follows:

\begin{equation}\label{q_17}
I({S_1},{S_2}) = \frac{{\sum\limits_{i = {S_1}(1)}^{\left| {{S_1}} \right|} {\sum\limits_{j = {S_2}(1)}^{\left| {{S_2}} \right|} {{E_{ij}}} } }}{{\left| {{S_1}} \right| \cdot \left| {{S_2}} \right|}}
\end{equation}

where $i \ne j$, $S_1,S_2 \in {2^\Omega }$. ${S_1}(1)$ represent the first element in the set $S_1$, so do the set ${\left| {{S_2}} \right|}$. ${\left| {{S_1}} \right|}$ denotes the cardinality of $S_1$, and ${\left| {{S_2}} \right|}$ denotes the cardinality of $S_2$. Note that, when $i = j$, $I=1$.
\end{definition}

\subsection{Distance between two D numbers}

Since D numbers theory is a generalization of the Dempster-Shafer theory, a body of D numbers can be seen as a discrete random variable whose values are ${2^\Omega }$ with a probability distribution $d$. Based on above, a D number is a vector $\vec d$ of the vector space. Thus, the distance between two D numbers is defined as follows.

\begin{definition}
Let $d_1$ and $d_2$ be two D numbers on the same frame of discernment $\Omega$, containing N elements which are not require to be exclusive with each other. The distance between $d_1$ and $d_2$ is:

\begin{equation}\label{q_17}
d_{D-number({d_1},{d_2})} = \sqrt {\frac{1}{2}{{(\vec {{d_1}}  - \vec {{d_2}} )}^T}\underline{\underline D} {\cdot}\underline{\underline  I} (\vec {{d_1}}  - \vec {{d_2}} )}
\end{equation}

where ${\underline{\underline D} }$ and ${\underline{\underline I} }$ are two $({2^\Omega } \times {2^\Omega })$-dimensional matrixes. The elements of ${\underline{\underline D} }$ are:

$D(A,B) = \frac{{\left| {{A} \cap {B}} \right|}}{{\left| {{A} \cup {B}} \right|}}$. $(A,B \in {2^\Omega })$.

The elements of ${\underline{\underline I} }$ are:

$I({A},{B}) = \frac{{\sum\limits_{i = {A}(1)}^n {\sum\limits_{j = {B}(1)}^m {{E_{ij}}} } }}{{\left| {{A}} \right| \cdot \left| {{B}} \right|}}$. $(i \ne j)$, $(A,B \in {2^\Omega })$, (when $i = j, I=1$).

\end{definition}

An illustrative example is given to show the calculation of the distance between two D numbers step by step.

\begin{figure}[htbp]
\begin{center}
\psfig{file=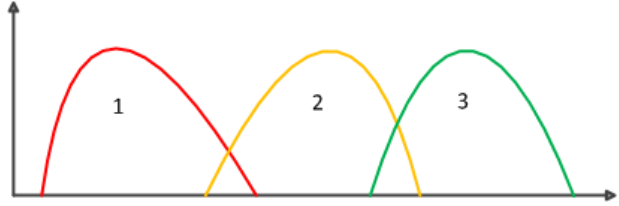,scale=0.75} \caption{Three contents which are not exclusive} \label{example2}
\end{center}
\end{figure}

\textbf{Example 2.}
Let $\Omega$ be a frame of discernment with 3 elements. We use 1,2,3 to denote element 1, element 2, and element 3 in the frame. The relationship of the three elements are shown in Fig. \ref{example2}.

Given ${U_{12}}=2$, ${U_{23}}=3$, ${S_{12}}=0.2$, ${S_{23}}=0.6$. Two bodies of D numbers are given as follows:

${d_1}(\{ 1\} ) = 0.3$, ${d_1}(\{ 1,2\} ) = 0.4$, ${d_1}(\{ 1,2,3\} ) = 0.4$.

${d_2}(\{ 2\} ) = 0.2$, ${d_2}(\{ 2,3\} ) = 0.3$, ${d_2}(\{ 1,2,3\} ) = 0.5$.

\textbf{Step 1} Constructing the vector $\vec {{d_1}}$ and $\vec {{d_2}}$:

$\vec {{d_1}}  = \left( \begin{array}{c}
0.3\\
0\\
0\\
0.4\\
0.3
\end{array} \right)$, $\vec {{d_2}}  = \left( \begin{array}{c}
0\\
0.2\\
0.3\\
0\\
0.5
\end{array} \right)$.

${\vec {{d_1}}  - \vec {{d_2}} }=\left( \begin{array}{c}
0.3\\
-0.2\\
-0.3\\
0.4\\
-0.2
\end{array} \right)$

\textbf{Step 2} Based on the given $U$ and $S$, the relative matrix can be calculated as below:

  $R=
\left[
\begin{array}{ccccccc}
      &  1     &  2    & 3\\
 1  &    1     &  0.1  & 0  \\
 2  &    0.1   &  1    & 0.2  \\
 3  &    0     &  0.2  & 1
\end{array}
\right]$

\textbf{Step 3} Calculate the distance matrix $\underline{\underline  D}$ according to Eq. \ref{q_9}. For example, the distance between ${1}$ and ${1,2}$ can be calculated as follows:

$D(\{ 1\} ,\{ 1,2\} ) = \frac{{\left| {\{ 1\} } \right|}}{{\left| {\{ 1,2\} } \right|}} = \frac{1}{2}$

The distance matrix $\underline{\underline  D}$ is:

  $\underline{\underline  D}=
\left(
\begin{array}{ccccccc}
 1  &   0   &  0  & 1/2  & 1/3 \\
 0  &   1   &  0  & 1/2  & 1/3  \\
 0  &   0   &  1  & 0    & 1/3  \\
 1/2  &   1/2  &  0  & 1  & 2/3 \\
 1/3  &1/3  &1/3&2/3&1
\end{array}
\right)$

\textbf{Step 4} Calculate the intersection matrix $\underline{\underline  I}$ according to Eq. \ref{q_17}. For example, the intersection degree between ${1,2}$ and ${1,2,3}$ can be calculated as follows:

$n={\left| {{S_1}} \right|}=2$, $m={\left| {{S_2}} \right|}=3$.

$I({S_1},{S_2}) = \frac{{\sum\limits_{i = {S_1}(1)}^n {\sum\limits_{j = {S_2}(1)}^m {{E_{ij}}} } }}{{\left| {{S_1}} \right| \cdot \left| {{S_2}} \right|}}
=\frac{{\sum\limits_{i = 1}^2 {\sum\limits_{j = 1}^3 {{E_{ij}}} } }}{{2 \cdot 3}}
=\frac{{0.1 + 0 + 0.1 + 0.2}}{6}=0.0667$

The intersection matrix is built as follows:

  $\underline{\underline  I}=
\left(
\begin{array}{ccccccc}
 1  &   0.1   &  0  & 0.05  & 0.0333 \\
 0.1  &   1   &  0.2  & 0.05  & 0.1  \\
 0  &   0.2   &  1  & 0.1    & 0.0667  \\
 0.05  &   0.05  &  0.1  & 1  & 0.0667 \\
 0.0333  &0.1  & 0.0667  & 0.0667  &1
\end{array}
\right)$

\textbf{Step 5} Calculate the distance between the two D numbers according to Eq. \ref{q_17}.

$d_{D-number({d_1},{d_2})} = \sqrt {\frac{1}{2}{{(\vec {{d_1}}  - \vec {{d_2}} )}^T}\underline{\underline D} {\cdot}\underline{\underline  I} (\vec {{d_1}}  - \vec {{d_2}} )}
=0.4312$

\section{Case study}\label{Case study}
In this section, some examples are given to study the discipline of distance between two D numbers. The following example shows the difference between $d_{BPA}$ and $d_{D-number}$ in an extreme situation.

\begin{figure}[htbp]
\begin{center}
\psfig{file=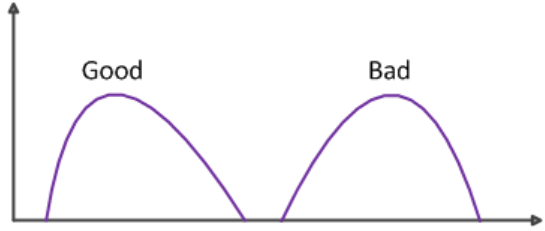,scale=0.75} \caption{The relationship between two linguistic constants in situation one.} \label{example31}
\end{center}
\end{figure}

\textbf{Example 3.} Let $\Omega$ be a frame of discernment with 2 linguistic constants, namely $\Omega  = \{ Good, Bad\}$. The relationship between the two linguistic constants is shown in Fig. \ref{example31}.

As the Fig. \ref{example31} shows, two constants are mutually exclusive. The distance function of D numbers is reasonable only if the discernment elements in the framework are not mutually exclusive, so we use the distance between two BPAs to calculate the distance of the two linguistic constants in Fig. \ref{example31}.

Given two pairs of BPAs:
${m_1}(\{ Good\} ) = 1$, ${m_1}(\{ Bad\} ) = 0$; ${m_2}(\{ Good\} ) = 0$, ${m_2}(\{ Bad\} ) = 1$.

The distance between the two BPAs can be calculated as bellow:

${d_{BPA}} = \sqrt {\frac{1}{2}(1, - 1)\left( \begin{array}{c}
1,0\\
0,1
\end{array} \right)\left( \begin{array}{c}
1\\
 - 1
\end{array} \right)}  = 1$

\begin{figure}[htbp]
\begin{center}
\psfig{file=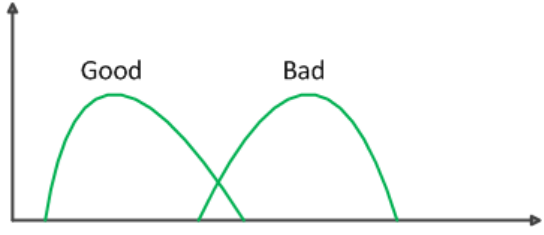,scale=0.75} \caption{The relationship between two linguistic constants in situation two.} \label{example32}
\end{center}
\end{figure}

If the relationship between the two linguistic constants are shown in the Fig. \ref{example32}, the two linguistic constants are not exclusive. The distance between two BPAs can not be used in this situation, so we use the proposed distance function of D numbers to calculate it.

Given two pairs of D numbers:
${d_1}(\{ Good\} ) = 1$, ${d_1}(\{ Bad\} ) = 0$; ${d_2}(\{ Good\} ) = 0$, ${d_2}(\{ Bad\} ) = 1$.

Assume that the relationship between the two D numbers can be expressed as ${E_{12}} = 0.2$. The distance between the two D numbers can be calculated as bellow:

${d_{D - num({d_1},{d_2})}} = \sqrt {\frac{1}{2}\left( {1, - 1} \right)\left( \begin{array}{l}
1,0\\
0,1
\end{array} \right)\left( \begin{array}{c}
1,0.2\\
0.2,1
\end{array} \right)\left( \begin{array}{c}
1\\
 - 1
\end{array} \right)}  = \sqrt {0.8}  = 0.8944$

From the example above, the proposed distance function of D numbers is reasonable when the discernment elements in the framework are not mutually exclusive. In the situation that the discernment is mutually exclusive, the proposed distance function of D numbers is degenerated as the distance function between two BPAs. This example proved that the proposed distance function based on D numbers is reasonable and effective. Another two examples are given to compare the difference between the $d_{BPA}$ and the $d_{D-number}$ as follows.

\textbf{Example 4.}
Let $\Omega$ be a frame of discernment with 20 elements. We use 1,2,etc. to denote element 1, element 2, etc. The elements in the frame of discernment are exclusive between each other. In this situation, given two pairs of BPAs and D numbers as follows:

${m_1}(\{ 2,3,4\} ) = 0.3$, ${m_1}(\{ 7\} ) = 0.4$, ${m_1}(\{ \Omega\} ) = 0.4$, ${m_1}(\{ A\} ) = 0.8$.

${m_2}(\{ 1,2,3,4,5\} ) = 1$.

${d_1}(\{ 2,3,4\} ) = 0.3$, ${d_1}(\{ 7\} ) = 0.4$, ${d_1}(\{ \Omega\} ) = 0.4$, ${d_1}(\{ A\} ) = 0.8$.

${d_2}(\{ 1,2,3,4,5\} ) = 1$.

where $A$ is a subset of $\Omega$.

The relative matrix can be calculated as below:

  $R=
\left[
\begin{array}{ccccccc}
      &  1     &  2    & 3  & 4  &\ldots  &20 \\
 1  &    1     &  0  & 0  & 0  & \ldots  & 0\\
 2  &    0   &  1    & 0& 0  & \ldots  & 0  \\
 3  &    0     &  0  & 1& 0  & \ldots  &0   \\
  4  &    0     &  0  & 0& 1  & \ldots  &0   \\
  \vdots  &   \vdots   & \vdots  & \vdots& \vdots& \vdots &\vdots \\
   20 &    0     &  0  & 0& 0  & \ldots  &1  \\
\end{array}
\right]$

\begin{figure}[htbp]
\begin{center}
\psfig{file=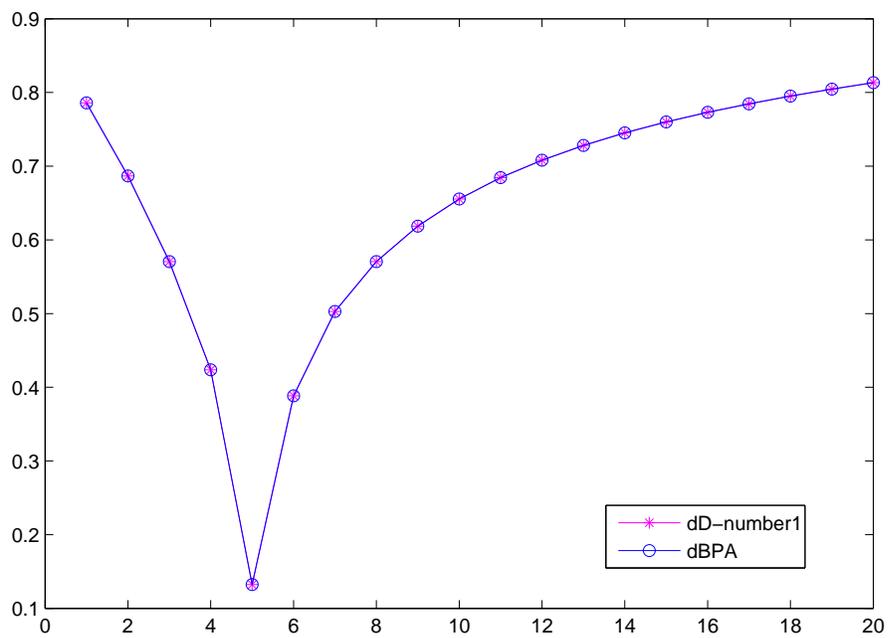,scale=0.75} \caption{The difference between $d_{BPA}$ and $d_{D-number1}$ in mutually exclusive situation.} \label{case1}
\end{center}
\end{figure}

There are 20 cases where subset $A$ increments one or more element at a time, starting from Case 1 with $A={1}$ and ending with Case 20 when $A=\Omega$. The distance between two BPAs $d_{BPA}$ defined in \cite{jousselme2001new} and the distance between two D numbers $d_{D-number1}$ proposed in this paper in this condition are shown in the Table \ref{tab_1} and graphically illustrated in Fig. \ref{case1}

The Fig. \ref{case1} shows that in the situation that the discernment is mutually exclusive, the proposed distance function of D numbers $d_{D-number1}$ is degenerated as the distance function between two BPAs $d_{BPA}$.

\begin{figure}[htbp]
\begin{center}
\psfig{file=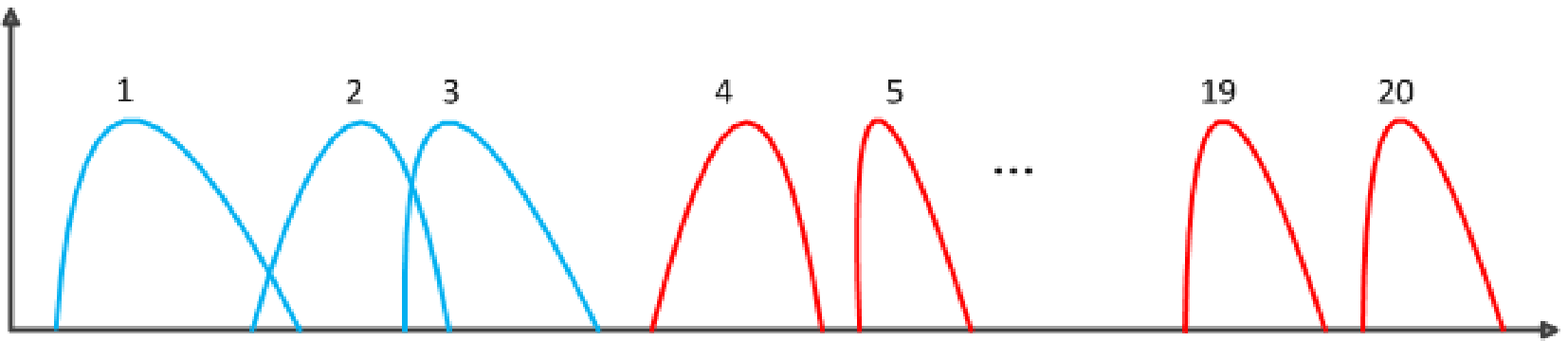,scale=0.75} \caption{Relationship between each two elements in the discernment} \label{case}
\end{center}
\end{figure}

\textbf{Example 5.}
Let $\Omega$ be a frame of discernment with 20 elements. We use 1,2,etc. to denote element 1, element 2, etc. in the frame. The relationship between each two elements is shown in Fig. \ref{case}. In this case, let element 4 to element 20 be exclusive from each other, and given ${U_{12}}=2$, ${U_{23}}=3$, ${S_{12}}=0.2$, ${S_{23}}=0.6$.

Based on the given $U$ and $S$, the relative matrix can be calculated as below:

  $R=
\left[
\begin{array}{ccccccc}
      &  1     &  2    & 3  & 4  &\ldots  &20 \\
 1  &    1     &  0.1  & 0  & 0  & \ldots  & 0\\
 2  &    0.1   &  1    & 0.2& 0  & \ldots  & 0  \\
 3  &    0     &  0.2  & 1& 0  & \ldots  &0   \\
  4  &    0     &  0  & 0& 1  & \ldots  &0   \\
  \vdots  &   \vdots   & \vdots  & \vdots& \vdots& \vdots &\vdots \\
   20 &    0     &  0  & 0& 0  & \ldots  &1  \\
\end{array}
\right]$

The first D number ${d_1}$ is defined as

${d_1}(\{ 2,3,4\} ) = 0.3$, ${d_1}(\{ 7\} ) = 0.4$, ${d_1}(\{ \Omega\} ) = 0.4$, ${d_1}(\{ A\} ) = 0.8$.

where $A$ is a subset of $\Omega$. The second D number ${d_2}$ used in the example is

${d_2}(\{ 1,2,3,4,5\} ) = 1$.

There are 20 cases where subset $A$ increments one or more element at a time, starting from Case 1 with $A={1}$ and ending with Case 20 when $A=\Omega$ as shown in Table \ref{tab_1}. The distance between two D numbers $d_{D-number2}$ in this situation for these 20 cases is detailed in Table \ref{tab_1} and graphically illustrated in Fig. \ref{compare}.

\begin{table}[htbp]
\caption{Comparison of $d_{BPA}$ and $d_{D-number}$ values of $d_1$ and $d_2$ when subset $A$ changes.}
\label{tab_1}
\begin{center}
\begin{tabular}{lcccccccccccccc}
\toprule Cases   &   &   &   & $d_{BPA}$&   &   &   &  $d_{D-number1}$&   &   &   &  $d_{D-number2}$ \\
\midrule
A = \{ 1\}       &   &   &   &  0.7858  &   &   &   &  0.7858&   &   &   &  0.7788     \\
A = \{ 1,2\}     &   &   &   &  0.6867  &   &   &   &  0.6867&   &   &   &  0.6721     \\
A = \{ 1,2,3\}   &   &   &   &  0.5705  &   &   &   &  0.5705&   &   &   &  0.5589     \\
A = \{ 1,...,4\} &   &   &   &  0.4237  &   &   &   &  0.4237&   &   &   &  0.4180     \\
A = \{ 1,...,5\} &   &   &   &  0.1323  &   &   &   &  0.1323&   &   &   &  0.1322     \\
A = \{ 1,...,6\} &   &   &   &  0.3884  &   &   &   &  0.3884&   &   &   &  0.3857     \\
A = \{ 1,...,7\} &   &   &   &  0.5029  &   &   &   &  0.5029&   &   &   &  0.4999     \\
A = \{ 1,...,8\} &   &   &   &  0.5705  &   &   &   &  0.5705&   &   &   &  0.5677     \\
A = \{ 1,...,9\} &   &   &   &  0.6187  &   &   &   &  0.6187&   &   &   &  0.6162     \\
A = \{ 1,...,10\}&   &   &   &  0.6554  &   &   &   &  0.6554&   &   &   &  0.6532      \\
A = \{ 1,...,11\}&   &   &   &  0.6844  &   &   &   &  0.6844&   &   &   &  0.6826      \\
A = \{ 1,...,12\}&   &   &   &  0.7081  &   &   &   &  0.7081&   &   &   &  0.7066      \\
A = \{ 1,...,13\}&   &   &   &  0.7281  &   &   &   &  0.7281&   &   &   &  0.7268      \\
A = \{ 1,...,14\}&   &   &   &  0.7451  &   &   &   &  0.7451&   &   &   &  0.7440      \\
A = \{ 1,...,15\}&   &   &   &  0.7600  &   &   &   &  0.7600&   &   &   &  0.7590      \\
A = \{ 1,...,16\}&   &   &   &  0.7730  &   &   &   &  0.7730&   &   &   &  0.7722      \\
A = \{ 1,...,17\}&   &   &   &  0.7846  &   &   &   &  0.7846&   &   &   &  0.7840      \\
A = \{ 1,...,18\}&   &   &   &  0.7951  &   &   &   &  0.7951&   &   &   &  0.7945      \\
A = \{ 1,...,19\}&   &   &   &  0.8046  &   &   &   &  0.8046&   &   &   &  0.8042      \\
A = \{ 1,...,20\}&   &   &   &  0.8133  &   &   &   &  0.8133&   &   &   &  0.8139      \\
\bottomrule
\end{tabular}
\end{center}
\end{table}

\begin{figure}[htbp]
\begin{center}
\psfig{file=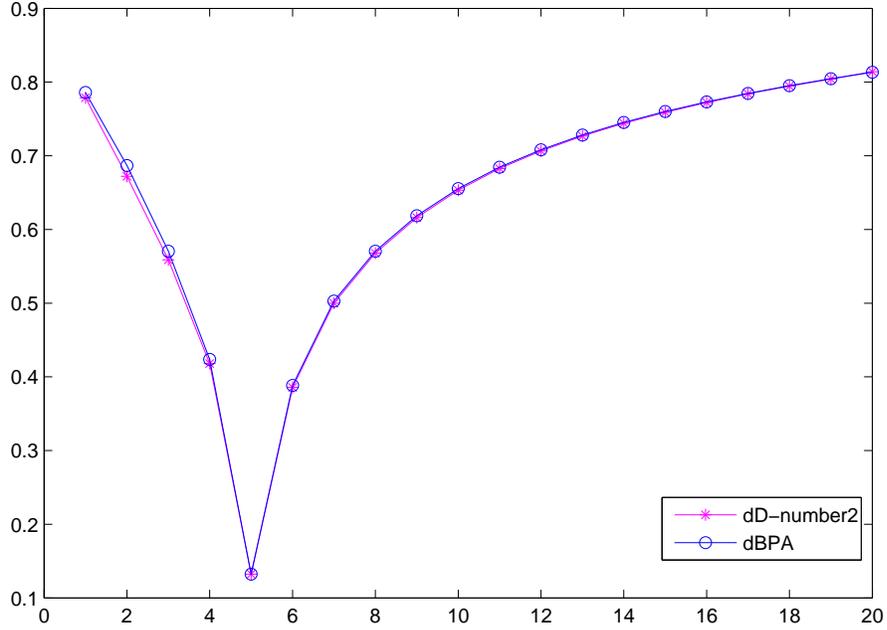,scale=0.75} \caption{The difference between $d_{BPA}$ and $d_{D-number2}$ when the elements of the frame are not mutually exclusive.} \label{case2}
\end{center}
\end{figure}

From the Table \ref{tab_1} and Fig. \ref{case2}, both measures go up and down consistently, when the size of $A$ changes. However, the $d_{D-number}$ values are always little smaller than the corresponding $d_{BPA}$ values, because the elements in set $\Omega$ do not require mutually exclusive in D numbers. The Example 4 and Example 5 proved that the proposed distance function of D numbers is effective when the elements in the frame of discernment are not mutually exclusive. When the discernment is mutually exclusive, the distance function of D numbers is degenerated as the distance function between two BPAs defined by Anne-Laure Jousselme.

\section{Conclusion}\label{Conclusion}
In this paper, a new distance function to measure the distance between two D numbers is proposed. D number is a new representation of uncertain information, which inherits the advantage of the Dempster-Shafer theory of evidence and overcomes some shortcomings. The proposed distance function of D numbers is effective when the elements in the frame of discernment are not mutually exclusive. In the situation that the elements in the frame of discernment are mutually exclusive, the proposed distance function of D numbers is degenerated as the distance function between two BPAs defined by Anne-Laure Jousselme.

\section*{Acknowledgements}
The work is partially supported by National Natural Science Foundation of China (Grant No. 61174022), Specialized Research Fund for the Doctoral Program of Higher Education (Grant No. 20131102130002), R\&D Program of China (2012BAH07B01), National High Technology Research and Development Program of China (863 Program) (Grant No. 2013AA013801), the open funding project of State Key Laboratory of Virtual Reality Technology and Systems, Beihang University (Grant No.BUAA-VR-14KF-02).

%



\bibliographystyle{elsarticle-num}
\bibliography{reference}







\end{document}